\pdfoutput=1
\documentclass[11pt]{article}
\newif\ifsubmission
\submissiontrue
\usepackage{acl}

\usepackage[T1]{fontenc}
\usepackage[utf8]{inputenc}
\usepackage{times}
\usepackage{latexsym}
\usepackage{microtype}
\usepackage{inconsolata}
\usepackage{graphicx}
\usepackage{booktabs}
\usepackage{multirow}
\usepackage{tabularx}
\usepackage{amsmath}
\usepackage{amssymb}
\usepackage{url}
\usepackage{xcolor}

\newif\ifshowmarkup
\showmarkupfalse   

\ifshowmarkup
  \newcommand{\nm}[1]{\textcolor{blue}{#1}}
  \newcommand{\risk}[1]{\textcolor{red}{#1}}
\else
  \newcommand{\nm}[1]{#1}
  \newcommand{\risk}[1]{#1}
\fi

\title{Rate-Resolved Text Evaluation with Masked Diffusion Language Models}

\author{
  Yiqi Liu$^{1}$ \quad
  Kun Zhao$^{2}$ \quad
  Chenghua Lin$^{1}$ \quad
  Nafise Sadat Moosavi$^{3}$ \\[2pt]
  $^{1}$The University of Manchester \quad
  $^{2}$University of Pittsburgh \quad
  $^{3}$The University of Sheffield \\[2pt]
  \texttt{\{yiqi.liu,chenghua.lin\}@manchester.ac.uk} \\
  \texttt{kun.zhao@pitt.edu}
  \quad
  \texttt{n.s.moosavi@sheffield.ac.uk}
}

\begin{document}

\maketitle

\begin{abstract}
\nm{Text evaluation is increasingly driven by scalar judgments from autoregressive likelihoods, prompted judges, and learned evaluators. While effective for ranking systems, scalar scores collapse diverse evaluation behavior into a single dimension. We propose rate-resolved evaluation, a reconstruction-based evaluation paradigm using masked diffusion language models. Masked diffusion models reconstruct text under varying masking rates, naturally exposing how quality signals behave as contextual evidence decreases. We preserve this reconstruction structure as a reconstruction profile: a vector of band-conditioned losses that captures how different quality signals persist or degrade across context regimes.
 On SummEval, preserving masking-rate structure provides complementary information beyond the scalar baselines tested here, improving fusion with autoregressive likelihood and prompted judges. On WMT MQM zh-en, the same reconstruction process produces different evaluation behaviors at different masking rates, yielding a tunable tradeoff between ranking translation quality and localizing error spans.
Our results suggest that, in the tested settings, diffusion reconstruction can expose evaluative structure that is not captured by the scalar baselines we evaluate, through structured recoverability under controlled corruption.
}
\end{abstract}
\section{Introduction}

\begin{figure*}[t]
  \centering
  \includegraphics[width=\textwidth]{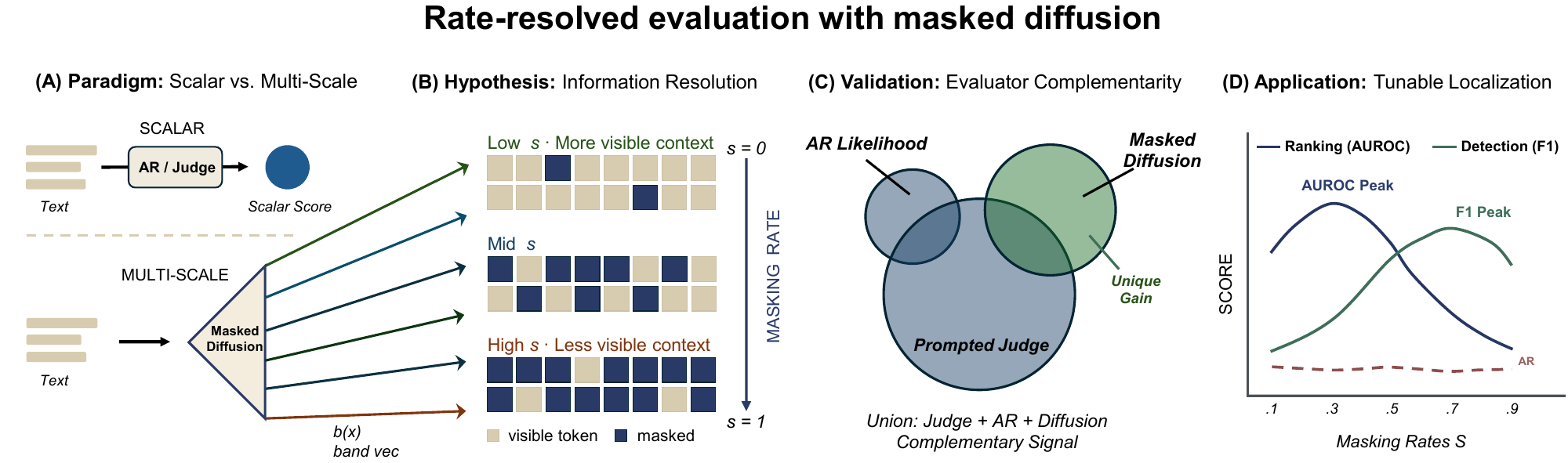}
  \caption{
  Instead of reducing text evaluation to a single scalar score, masked diffusion reconstructs text under different masking rates and yields a band-conditioned score vector $\mathbf{b}(x)$. Lower masking rates retain more visible context, while higher masking rates require reconstruction from sparser evidence. The resulting masking-rate structure exposes complementary evaluation behavior beyond scalar evaluation.
  }
  \label{fig:overview}
\end{figure*}

\nm{A single evaluation score is often enough to rank generation systems, but it compresses diverse evaluation behavior into one dimension. Two summaries may receive similar scores while differing substantially in coherence, contextual dependence, or sensitivity to localized errors. Existing evaluators, including autoregressive likelihoods, learned metrics, and prompted judges, primarily produce scalar judgments that summarize model preference or confidence \cite{yuan2021bartscoreevaluatinggeneratedtext, qin2022t5scorediscriminativefinetuninggenerative, liu-etal-2023-g, fu-etal-2024-gptscore, zhong-etal-2022-towards}. While effective for ranking systems, scalar evaluation can obscure complementary structure in how text quality behaves under different levels of contextual evidence.}

\nm{
In this paper, we investigate masked diffusion language models as a reconstruction-based perspective on text evaluation. These models reconstruct text under varying masking rates, requiring reconstruction from progressively reduced context. This reconstruction process naturally exposes how quality signals behave as contextual evidence decreases. We call this view \emph{rate-resolved evaluation}: instead of reducing evaluation to a single scalar judgment, we preserve masking rate as a model-native evaluation axis and represent each text as a reconstruction profile across masking regimes.
}

\nm{The resulting reconstruction profile is indexed by masking rate, a variable already present in the diffusion objective. Different masking regimes probe text quality under different levels of contextual evidence, revealing evaluation behaviors that scalar metrics can obscure. For example, some masking regimes better capture summary-level quality differences, while others are more sensitive to localized errors. This creates a structured evaluation space that can expose complementary signals for ranking and error localization. Unlike prompt-based decompositions or manually engineered evaluation features, this structure arises directly from the evaluator’s reconstruction process itself.
}

\nm{Figure~\ref{fig:overview} illustrates this paradigm. Rather than assigning a single evaluation score to a text, masked diffusion produces a vector of band-conditioned reconstruction losses across masking regimes. We investigate whether this masking-rate structure provides complementary evaluative information beyond scalar metrics and judges, and whether different masking regimes expose different evaluation behaviors. In this paper, we study these questions through two representative settings: summary-level evaluation, which measures overall text quality for system ranking, and token-level localization, which identifies where errors occur within a text.
}

\nm{The paper makes three \textbf{main contributions}. First, we show how masked diffusion language models naturally support rate-resolved evaluation through reconstruction under varying masking rates, turning masking rate into a structured evaluation axis for text quality assessment. Second, we show that different masking regimes expose complementary evaluative information beyond scalar metrics and prompted judges on summary evaluation benchmarks. Third, we show that different masking regimes support different evaluation objectives, with some better suited for ranking text quality and others more effective for localizing errors.
}

\section{Related Work}

\nm{We organize prior work by how evaluators represent text quality: scalar evaluators, externally designed evaluation axes, and model-native evaluation structure.}

\paragraph{Scalar evaluators.}
Existing text-generation metrics range from lexical overlap to embedding-based,
likelihood-based, trained, and prompted evaluators
\cite{papineni-etal-2002-bleu, lin-2004-rouge,
zhang2020bertscoreevaluatingtextgeneration, zhao-etal-2019-moverscore,
sellam-etal-2020-bleurt, yuan2021bartscoreevaluatinggeneratedtext,
qin2022t5scorediscriminativefinetuninggenerative, fu-etal-2024-gptscore,
zhong-etal-2022-towards}. Prompted judge models have become especially strong:
G-Eval and related systems can correlate closely with human judgments in many
settings
\cite{liu-etal-2023-g, wang-etal-2023-chatgpt, chiang-lee-2023-large}. 
\nm{Despite their methodological differences, most existing evaluators ultimately map each text to a single scalar judgment. Our work instead studies whether evaluation can be represented as structured behavior across masking regimes, exposing complementary signals that scalar evaluation does not capture.}

\paragraph{Masked-LM and cloze-style evaluation.}
Cloze-style and masked-LM evaluators also use reconstruction as an evaluation
signal. BLANC estimates summary quality by measuring how much access to a
summary helps a pretrained language model solve cloze-style prediction tasks on
the source document \cite{vasilyev-etal-2020-fill}. Masked-LM scoring uses
pseudo-log-likelihoods computed by masking tokens one at a time
\cite{salazar-etal-2020-masked}. Our work differs by preserving the
masking-rate axis of a masked diffusion evaluator: rather than reducing
reconstruction to a single cloze or pseudo-likelihood score, we study how
evaluation behavior changes across masking regimes.

\paragraph{Externally designed evaluation structure.}
\nm{Behavioral test suites, contrast sets, and dynamic challenge benchmarks expose evaluation behaviors that aggregate scores can hide
\cite{ribeiro-etal-2020-beyond, gardner-etal-2020-evaluating, kiela-etal-2021-dynabench}. In these approaches, the analyst or data-collection process specifies controlled evaluation conditions through templates, perturbations, contrast examples, or adversarially collected cases. Such methods provide structured views of model behavior, but the structure is imposed by the evaluation design rather than arising from the evaluator’s own modeling objective.}

\paragraph{Model-native evaluation structure in diffusion models.}
\nm{Text diffusion models have primarily been developed for generation. Important examples include Diffusion-LM \cite{li2022diffusionlmimprovescontrollabletext},
DiffuSeq \cite{gong2023diffuseqsequencesequencetext},
self-conditioned embedding diffusion \cite{strudel2022selfconditionedembeddingdiffusiontext},
AR-Diffusion \cite{wu2023ardiffusionautoregressivediffusionmodel},
and later work on bridging the training and inference gap
\cite{tang-etal-2023-diffusion}.
 A
common feature of these models is iterative reconstruction under a noise or
masking schedule, where generation behavior changes as contextual evidence is
progressively removed or restored. Prior work has largely used this structure
for controllable or higher-quality generation. We instead investigate masking
rate as a model-native evaluation axis, asking whether reconstruction behavior
across masking regimes exposes complementary evaluative information beyond
scalar metrics and judges.}
\section{Rate-Resolved Evaluation}
\nm{Before introducing the formalism, we describe the core idea intuitively. Given a text, we choose a masking rate $s$, hide that fraction of tokens, ask the diffusion model to reconstruct the masked content, and measure the resulting reconstruction loss. Varying $s$ changes how much contextual evidence remains available to the model. Rather than aggregating reconstruction behavior across all masking rates into a single score, we preserve reconstruction behavior across masking regimes as a structured evaluation signal.}

\nm{The masking-rate axis can support different forms of evaluation behavior depending on how reconstruction signals are summarized or probed. In this paper, we study two complementary settings. For summary-level evaluation, we represent reconstruction behavior through masking-rate bands. For token-level localization, we measure how often tokens change under repeated mask-and-reconstruct revisions. Together, these settings illustrate how the same reconstruction process can support different evaluation objectives, including ranking text quality and localizing errors.
}

\subsection{Masking-Rate Reconstruction Profiles}
\nm{We now formalize masking-rate reconstruction profiles. Given a text, we sample a masking rate $s$, mask that fraction of tokens, ask the diffusion model to reconstruct the masked content, and measure the resulting reconstruction loss.
}

Let $x = (x_1, \ldots, x_N)$ be a token sequence and
let $m \sim \mathrm{Bern}(s)^N$ denote an i.i.d.\ masking pattern at masking
rate $s \in [0,1)$. We write $S = \{i : m_i = 1\}$ for the masked positions and
$\bar{S}$ for the visible positions. A masked diffusion language model
$p_\theta$ predicts each $x_i$ for $i \in S$ from the partially observed input
$x_{\bar{S}}$.

For a fixed masking rate $s$, we define the normalized masking-rate loss density
\begin{equation}
\ell(s, x) \triangleq
\mathbb{E}_{m}
\left[
\frac{1}{|S|}
\sum_{i \in S} r_i(x,m)
\;\middle|\; |S| > 0
\right],
\label{eq:spectral-density}
\end{equation}
where $r_i(x,m)=-\log p_\theta(x_i \mid x_{\bar{S}})$.
This quantity measures the average masked-token reconstruction loss under a random
mask pattern at rate $s$. Higher loss means that the observed text is harder
for the model to reconstruct under that masking regime. 
\nm{Because masking patterns are
sampled randomly, the degenerate event $|S|=0$ can occasionally occur at low
masking rates; in Monte Carlo estimation, we simply discard these samples.}
When the evidence
lower bound (ELBO) is written in the equivalent $m_i r_i(x,m)$ form, $r_i$ is
only needed on masked positions and the product is zero for $i \notin S$.

\nm{This masking-rate-dependent reconstruction loss also connects naturally to the
diffusion training objective.} For the absorbing-mask family used by masked diffusion models, the training
objective can be written as a weighted integral over masking rates. 
\begin{equation}
\mathcal{L}_{\mathrm{diff}}(x)
\approx
\int_0^{1^-} \frac{s}{1-s}\, N \,\ell(s,x)\, ds,
\label{eq:diff-elbo}
\end{equation}

\nm{We never evaluate $s=1$ directly and sample only rates with $s<1$ in practice.
Appendix~\ref{app:derivation} gives the derivation and details the approximation
used to replace the random number of masked positions by its expectation.\footnote{
We verified on a SummEval subset that this approximation has negligible effect
on document-level rankings (Spearman(exact, approximate) $\geq 0.9969$ across
all tested rates).}
}

We then define the loss of a masking-rate band $[a,b) \subseteq [0,1)$ as
\begin{equation}
\mathcal{L}_{[a,b]}(x)
\triangleq
\int_a^b \frac{s}{1-s}\, N \,\ell(s,x)\, ds.
\label{eq:band-loss}
\end{equation}
In practice, we estimate Eq.~\ref{eq:band-loss} with Monte Carlo samples over
both masking rates and mask patterns. 
\nm{Each band summarizes reconstruction
behavior within a particular masking regime.}

We therefore represent each text with the masking-rate reconstruction profile
\begin{equation}
\mathbf{b}(x) =
\left(
\mathcal{L}_{[a_1,b_1]}(x), \ldots, \mathcal{L}_{[a_K,b_K]}(x)
\right),
\label{eq:band-vector}
\end{equation}
where the intervals form a partition of $[0,1)$. \nm{Each coordinate captures
reconstruction loss within a different range of masking rates, yielding a
structured evaluation profile rather than a single scalar score.\footnote{
In experiments, we use equal-width masking-rate bands, with the final band
covering rates close to $s=1$ while excluding the degenerate endpoint itself.
The derivation is specific to absorbing-mask diffusion models; extending the
same decomposition to other non-autoregressive generators would require
different assumptions.}}


\subsection{Interpreting the Masking-Rate Axis}
\nm{The masking-rate axis is operational rather than manually designed: each
coordinate in Eq.~\ref{eq:band-vector} corresponds to a reconstruction regime
already present in the diffusion objective. Different masking rates expose
different reconstruction behaviors because the amount of visible context changes
with $s$. For example, reconstruction under lower masking rates may better
reflect fine-grained consistency with nearby context, while higher masking rates
can become more sensitive to broader contextual coherence when local evidence is
sparse. In our experiments, different masking regimes favor different
evaluation behaviors, including summary-level ranking and localized error
detection.}

\nm{We therefore treat masking rate primarily as a structured evaluation axis rather
than assigning fixed semantic meanings to individual bands. While the precise
behavior of individual masking regimes may depend on the diffusion evaluator and
evaluation setting, the underlying masking-rate axis remains operationally
consistent: all generated texts are evaluated through the same family of
reconstruction conditions, yielding comparable reconstruction profiles across
systems and outputs. Unlike prompt-based decompositions or manually specified
evaluation dimensions, the masking-rate structure arises directly from the
evaluator’s reconstruction process itself.
}

\subsection{Autoregressive Likelihood as a Scalar Baseline}

For AR baselines, we use the standard full-sequence negative log-likelihood
\begin{equation}
\mathcal{L}_{\mathrm{AR}}(x)
=
-\sum_{i=1}^{N}\log p_\phi(x_i \mid x_{<i}),
\label{eq:ar-loss}
\end{equation}

\nm{which provides a single likelihood-based evaluation score for the entire text.
We compare this scalar baseline
with the masking-rate reconstruction profile in
Eq.~\ref{eq:band-vector} to
test whether different masking regimes provide complementary evaluative
information beyond standard autoregressive likelihood.}

\subsection{Roundtrip Reconstruction for Token-Level Localization}

\nm{The masking-rate axis can also be used for token-level evaluation. Instead of
measuring reconstruction difficulty for an entire text, we ask which tokens are
repeatedly revised under reconstruction. For a fixed masking rate $s$, we
repeatedly mask and reconstruct the same text under independently sampled mask
patterns.} Let $\hat{x}^{(k)}$ be the $k$-th reconstruction under mask pattern
$m^{(k)}$. We define the roundtrip conditional revision frequency for token $i$ as
\begin{equation}
\mathrm{RTP}_s(i,x)
\triangleq
\frac{
\sum_{k=1}^{K}
\mathbb{1}[m_i^{(k)} = 1]\,
\mathbb{1}[\hat{x}^{(k)}_i \neq x_i]
}{
\max\left\{1,\sum_{k=1}^{K}\mathbb{1}[m_i^{(k)} = 1]\right\}
}.
\label{eq:roundtrip}
\end{equation}
High values indicate tokens that are likely to change when exposed to
reconstruction. We normalize by the number of times a token is actually masked, so the score estimates conditional revision frequency rather than marginal exposure. Masking rate still controls the reconstruction context available to the model. We evaluate these token scores with AUROC and thresholded F1 against human span annotations.
\nm{Varying
$s$ therefore produces a localization operating curve across masking regimes,
allowing different masking rates to favor different localization behaviors. Unlike the band-conditioned losses, which are derived from the diffusion
objective, this score is a reconstruction-based heuristic designed for
token-level localization.}

\subsection{Evaluation Hypotheses}
\nm{The masking-rate framework suggests three empirical questions. (i) If reconstruction profiles capture evaluative structure beyond scalar
metrics, then preserving masking-rate information should provide complementary
evaluation behavior beyond standard scalar evaluators.}
\nm{(ii) If masking rate changes what reconstruction evidence is available to
the evaluator, then different masking regimes may support different forms of
token-level error localization. Lower masking rates may expose a broader set of
local inconsistencies, while higher masking rates may produce more selective
error signals under sparse context.}
\nm{(iii) if masking rate changes the reconstruction evidence available
to the evaluator, then the masking regimes that are most useful
should depend on the evaluation objective. We study this through
both summary-level evaluation and token-level localization.}
\section{Experimental Setup}

\paragraph{Benchmark roles.} The experiments assign each benchmark a specific role. 
\nm{\textbf{SummEval} \cite{fabbri-etal-2021-summeval} tests whether masking-rate
reconstruction profiles provide complementary evaluative information beyond AR
likelihood and prompted judges for multidimensional summary quality. \textbf{WMT MQM} zh-en, following the machine translation evaluation setting of
\citet{Kocmi2023LargeLM}, tests token-level localization behavior across
masking regimes. 
}

\nm{\paragraph{Datasets.} For WMT MQM, we use the released newstest2021 Chinese-to-English MQM TSV from the WMT21 metrics shared task and public MQM annotation release
\cite{freitag-etal-2021-results,google-wmt-mqm-human-evaluation}.
We retain the 8,431 segment-system hypotheses whose system id begins with
\texttt{hyp.}, covering 13 MT systems after excluding non-hypothesis entries.
Marked MQM spans are stripped from the target text and projected to
character-level gold masks. We evaluate all-error, major-error, and
minor-only severity slices.}

\begin{table}[t]
\centering
\small
\setlength{\tabcolsep}{5pt}
\begin{tabularx}{\linewidth}{@{}p{3.5cm}>{\raggedright\arraybackslash}X@{}}
\toprule
Role & Model \\
\midrule
Main diffusion & LLaDA-8B-Instruct \\
Replication diffusion & Dream-7B-Instruct \\
AR / judge & Qwen2.5-7B-Instruct \\
Static cloze control & XLM-R-XXL \\
\bottomrule
\end{tabularx}
\caption{Model roles in the main experiments. Exact checkpoints and inference
details are in Appendix~\ref{app:reproducibility}.}
\label{tab:model-roles}
\end{table}

\paragraph{Models.} The main diffusion models are LLaDA-8B-Instruct \cite{nie2025large} and
Dream-7B-Instruct \cite{dream2025dream}. Qwen2.5-7B-Instruct \cite{yang2025qwen25technicalreport} serves as both the AR
likelihood baseline and prompted judge.

\nm{\paragraph{Localization baseline.} We use XLM-R-XXL \cite{conneau-etal-2020-unsupervised} as a non-diffusion
masked-language-model baseline. Each token is independently masked and scored by reconstruction
cross-entropy under full left and right context. This tests whether a strong
static cloze-style evaluator can match the localization behavior of roundtrip
diffusion reconstruction.}

\paragraph{Evaluation protocols.} 
\nm{For summary-level evaluation, we report band-conditioned correlation analyses.
For localization, we report area under the receiver operating characteristic
curve (AUROC) and token-level F1. 
Fusion experiments use grouped cross-validation by source document, with
diffusion-band selection and linear fusion weights fit only on training folds.}


\paragraph{Implementation details.} Unless otherwise stated, band-loss experiments use 10 equal-width masking-rate
bands.  Roundtrip localization experiments use $s\in\{0.1,0.3,0.5,0.7,0.9\}$ with 100
reconstructions per rate.
\nm{Token-level F1 uses a fixed prediction threshold of 0.2 on the conditional
revision frequency in Eq.~\ref{eq:roundtrip}, chosen before test-set analysis. Appendix~\ref{app:reproducibility} gives inference settings, runtime notes,
fusion selection, and prompted-judge parsing details; Appendix
\ref{app:efficiency-sensitivity} reports Monte Carlo budget, band-boundary, and
threshold sensitivity controls.}

\section{Results}

\begin{figure*}[t]
  \centering
  \includegraphics[width=0.95\textwidth]{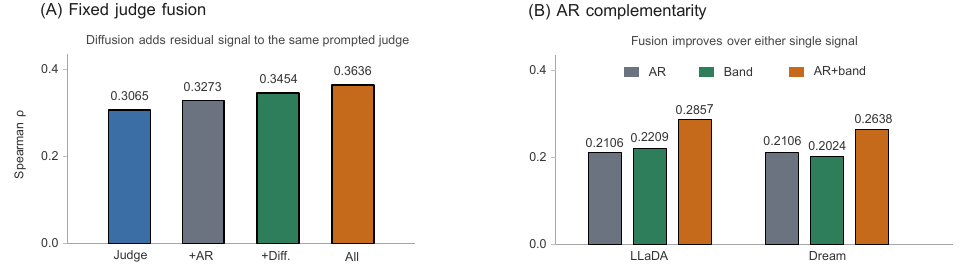}
  \caption{SummEval relevance complementarity. Left: LLaDA masking-rate
reconstruction bands remain complementary even to a strong prompted Qwen judge,
with judge+AR+diffusion achieving the strongest relevance correlation. Right:
for both LLaDA and Dream, combining AR likelihood with the best masking-rate
band consistently outperforms either signal alone, indicating that evaluative
signal is carried by reconstruction behavior across masking regimes rather than
by a single scalar diffusion score.
  }
  \label{fig:summeval-complementarity}
\end{figure*}

\subsection{Preserving Masking-Rate Structure Adds Evaluative Signal}

\nm{We first test the central evaluation hypothesis of the paper: whether preserving
masking-rate structure provides evaluative information beyond standard scalar
evaluators. SummEval provides a controlled setting for this comparison because
it contains multidimensional human quality judgments together with strong AR and
prompted-judge baselines. We evaluate this hypothesis in three steps: first by
testing complementarity with scalar evaluators, then through controls that
isolate masking-rate structure from generic diffusion features, and finally
through replication across diffusion evaluators.}

\paragraph{Complementarity with scalar evaluators.}
\nm{
The strongest and most consistent complementarity effect appears on the
relevance dimension. Figure~\ref{fig:summeval-complementarity} summarizes these
relevance results. For both LLaDA and Dream, combining AR likelihood with the
best masking-rate band substantially improves over either signal alone.
Smaller gains also appear for consistency and fluency, although the effect is
weaker and less consistent across models. Coherence is the main exception:
AR likelihood remains dominant there, and diffusion bands provide little
additional benefit. Table~\ref{tab:summeval-dimension-breakdown} in Appendix gives the full per-dimension breakdown.
}

\nm{
These gains persist even when the scalar evaluator itself is already strong.
Adding LLaDA diffusion reconstruction signals to a fixed prompted Qwen judge
improves relevance correlation from $0.3065$ to $0.3454$, while combining
judge, AR, and diffusion signals reaches the strongest overall performance at
\textbf{0.3636} (Table~\ref{tab:fixed-judge-fusion}; Figure~\ref{fig:summeval-complementarity}, left). The
judge+diffusion improvement over the judge alone remains statistically stable
under grouped bootstrap ($p=0.0080$), and the three-way
judge+AR+diffusion combination further improves over the judge baseline
($p<0.001$). 
}


\nm{These patterns suggest that masking-rate reconstruction may be especially
sensitive to evaluation dimensions that depend on broader semantic support.
High-mask reconstruction requires the evaluator to recover text under sparse
contextual evidence, which may make reconstruction difficulty more dependent
on whether summary content remains semantically recoverable from surrounding
context. In contrast, coherence and fluency are more closely tied to local sequential compatibility, where autoregressive likelihood already provides a strong signal.
Consistency lies between these extremes, producing smaller but still positive
gains.}

\begin{table}[t]
\centering
\small
\setlength{\tabcolsep}{5pt}
\begin{tabular}{@{}lc@{}}
\toprule
Evaluator & SummEval relevance $\rho$ \\
\midrule
Qwen judge & 0.3065 \\
Judge + AR & 0.3273 \\
Judge + diffusion & 0.3454 \\
Judge + AR + diffusion & 0.3636 \\
\bottomrule
\end{tabular}
\caption{Fixed judge fusion on SummEval relevance. Diffusion bands add
residual signal to the same prompted judge, and the strongest grouped-CV result
uses judge, AR, and diffusion together.}
\label{tab:fixed-judge-fusion}
\end{table}

\paragraph{The gain depends on preserving masking-rate structure.}
\nm{We next test whether the SummEval gains come specifically from preserving masking-rate structure rather than simply adding another diffusion-derived score. If the improvement came only from introducing an additional model signal, collapsing diffusion reconstruction into a single scalar should perform as well as the best masking-rate band. Appendix Table~\ref{tab:summeval-scalarization-controls} gives the full scalarization controls for LLaDA and Dream. On LLaDA relevance, the scalar control falls short: when fused with AR likelihood, the best masking-rate band reaches 0.2795, compared with 0.2610 for the corresponding full diffusion scalar ($\Delta=+0.0185$, $p=0.0015$). The same LLaDA comparison is positive on consistency (0.1705 vs.\ 0.1428, $\Delta=+0.0276$, $p=0.0085$). For Dream, the scalarization control is weaker and dimension-dependent, with a positive effect on consistency and little gain on relevance. These controls place the clearest evidence for masking-rate structure in LLaDA, with Dream providing a weaker replication.}

\nm{A complementary sample-shuffle control preserves the same 200 reconstruction
samples while destroying masking-rate order. Over 100 shuffles,
AR+pseudo-band fusion averages $\rho=0.2098$ on relevance and never matches
the true AR+rate-band result of $0.2795$. Together, these controls suggest
that evaluative signal is carried specifically by how reconstruction behavior
changes across masking regimes, rather than by arbitrary multidimensional
diffusion features.}

\paragraph{Replication across diffusion evaluators.}
\nm{To test whether this complementarity is specific to LLaDA, we repeat the
analysis with Dream-7B-Instruct. Dream reproduces the same qualitative pattern:
the best masking-rate band remains complementary to AR likelihood on relevance,
and AR+band fusion again outperforms either signal alone. The effect is weaker
for fluency and coherence, where single-band diffusion signals remain limited.
 Table~\ref{tab:summeval-dimension-breakdown} in Appendix gives the full Dream breakdown.}

\begin{figure*}[t]
  \centering
  \includegraphics[width=0.82\textwidth]{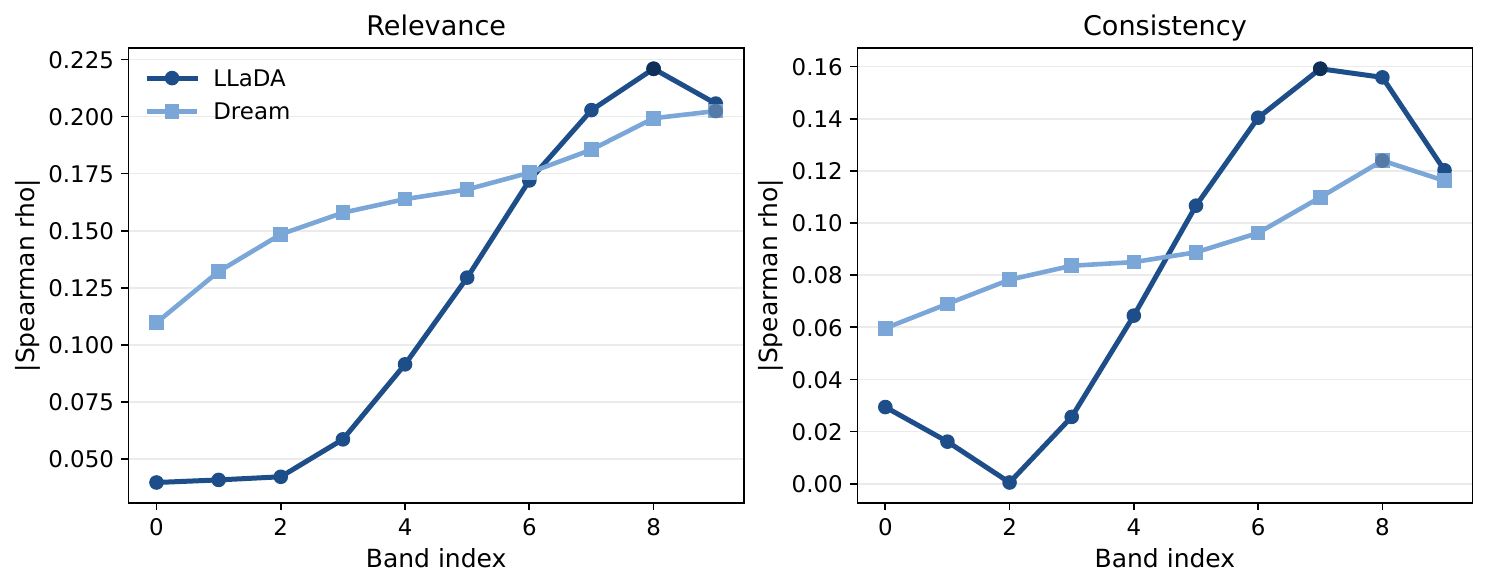}
  \caption{SummEval masking-rate profiles for relevance and consistency. Both
diffusion evaluators concentrate their strongest evaluative signal in high
masking-rate bands, although the peak location and magnitude differ across
models. Larger values indicate stronger alignment between diffusion
reconstruction loss and human quality judgments.
  }
  \label{fig:summeval-band-curves}
\end{figure*}

\paragraph{Masking-rate profiles.}
\nm{Figure~\ref{fig:summeval-band-curves} shows the corresponding masking-rate
profiles. Both diffusion evaluators concentrate their strongest relevance and
consistency signals in high masking-rate bands, although the peak location and
magnitude differ across models. Fusion gains are largest in the same regions
where diffusion bands diverge most strongly from AR rankings, suggesting that
high-mask reconstruction exposes evaluative behavior that scalar AR likelihood
does not fully capture. Appendix~\ref{app:inductive-bias-slices} gives
qualitative examples illustrating where judge, AR, and diffusion evaluators
fail differently.}

\subsection{Masking Rate Provides a Localization Operating Curve}
\label{sec:wmt-localization}

\begin{figure*}[t]
  \centering
  \includegraphics[width=0.88\textwidth]{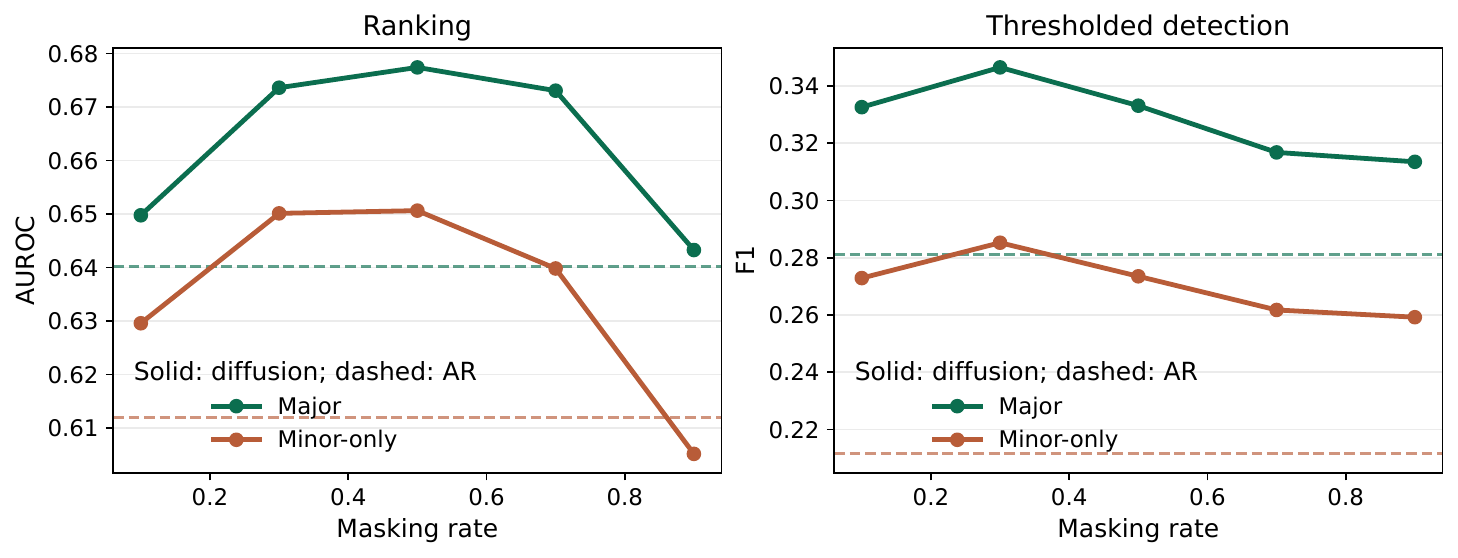}
  \caption{Masking-rate-dependent localization behavior on WMT MQM zh-en.
  Different masking regimes favor different localization objectives: intermediate
masking rates produce the strongest ranking behavior (AUROC), while lower
masking rates favor broader thresholded error detection (F1). Dashed horizontal
lines show AR token-surprisal baselines for the same severity slice.
  }
  \label{fig:wmtmqm-sigma}
\end{figure*}

\nm{We next ask whether masking rate exposes different localization behaviors. In
the SummEval experiments, masking-rate reconstruction profiles revealed
different summary-level evaluation signals. Here, we test whether the same
control axis also changes how the evaluator identifies localized translation
errors.}

\nm{WMT MQM provides a useful setting for this analysis because it contains dense
human error annotations at token-span level. Rather than assigning a single
quality score to an entire output, the evaluator must determine which regions
of the translation are likely to be problematic. We therefore use the
conditional roundtrip revision frequency from Eq.~\ref{eq:roundtrip} as a token-level
localization signal and examine how localization behavior changes across
masking regimes.}

\nm{From an evaluation perspective, the masking-rate axis defines a localization
operating curve rather than a single detection setting. Different masking
regimes may favor different evaluation objectives: some may better support
ranking suspicious regions, while others may favor broader thresholded error
coverage. The WMT experiments test whether these tradeoffs emerge
systematically across masking rates.}

\nm{Figure~\ref{fig:wmtmqm-sigma} shows that masking rate defines a clear
localization operating curve. Different masking regimes favor different
evaluation objectives. Ranking-oriented AUROC peaks near $s=0.5$, while
fixed-threshold F1 peaks earlier near $s=0.3$. The same qualitative pattern
appears across major and minor-only error slices, corresponding to severe
versus lower-severity MQM annotations.}

\nm{This separation is consistent with the interpretation that masking rate changes
the type of reconstruction evidence available to the evaluator. Intermediate
masking rates appear to produce more selective revision behavior, improving the
ability to rank suspicious regions reliably and yielding stronger AUROC. Lower masking rates retain denser local context, so when a token is selected for reconstruction, the model can make more locally grounded revisions. This produces broader thresholded error coverage and leads to higher F1. In this view, masking rate acts as an operational tradeoff between selective ranking behavior and broader error detection.}

\nm{The observed shifts are stable under bootstrap resampling. Overall AUROC peaks
at $0.6705$ near $s=0.5$, while overall F1 peaks at $0.2311$ near $s=0.3$.
The AUROC shift is small but consistent under 2,000 paired bootstrap resamples
($\Delta=0.0029$, 95\% CI $[0.0015,0.0044]$, $p=0.0005$), while the F1 shift is
larger ($\Delta=0.0127$, 95\% CI $[0.0108,0.0147]$, $p=0.0005$).}

\nm{Against AR token surprisal, roundtrip diffusion reconstruction is strongest for
ranking behavior. Overall AUROC reaches $0.6705$ at $s=0.5$, compared with
$0.6330$ for AR surprisal. The fixed-threshold F1 comparison is more mixed:
AR performs better on pooled overall F1, while diffusion performs better on
major and minor-only error slices. Under held-out-system threshold selection,
diffusion at $s=0.3$ also remains ahead of AR on macro-F1 for error-containing,
major-error, and minor-only slices
(Appendix~\ref{app:efficiency-sensitivity}).}

\nm{A bidirectional cloze baseline does not close the gap. XLM-R-XXL reaches
$0.6236$ overall AUROC, below both LLaDA roundtrip probing at $s=0.5$ and AR
surprisal. Although the comparison is not perfectly matched in model scale,
training data, or objective, the result suggests that the localization behavior
is not explained simply by access to bidirectional context. Appendix
\ref{app:dream-wmt} reports the weaker Dream replication, and Appendix
\ref{app:inductive-bias-slices} gives qualitative localization examples and
error-position stress tests.}

\section{Conclusion}
\nm{This work introduces masked diffusion language models as evaluators for text
generation. We show that diffusion reconstruction exposes useful evaluation
structure through the masking-rate axis: preserving reconstruction behavior
across masking regimes yields complementary information beyond autoregressive
likelihood and prompted judges, while also defining a localization operating
curve for token-level error detection.
The results suggest that, for the tested masked diffusion evaluators, evaluation behavior can depend not only on model scale or instruction tuning, but also on the reconstruction process used by the evaluator.
Reconstruction under partial observation probes generated text differently from
left-to-right prediction, and the resulting masking-rate profiles expose
evaluative structure not preserved in standard scalar metrics. In our
experiments, high masking rates are particularly informative for
relevance-oriented evaluation, while varying masking rate reveals systematic
tradeoffs between selective ranking behavior and broader localization coverage.
More broadly, modern evaluation increasingly relies on the same foundation
models used for generation itself. Prompted judges and autoregressive
likelihood evaluators often inherit closely related modeling assumptions and
inductive biases. Our findings suggest that diffusion models offer a distinct
evaluation perspective through iterative reconstruction under controlled
corruption.
Masking rate is one example of how model-native generation processes can expose
useful evaluation structure beyond single scalar scores. Future work may
identify additional evaluative behavior emerging from other reconstruction,
corruption, or sampling processes across different generative paradigms.
}
\section*{Limitations}

\paragraph{Model-family generalization.}
The current evidence is strongest for masked diffusion models, especially
LLaDA. Dream reproduces the SummEval relevance complementarity more weakly and
gives weaker WMT localization. The supported scope is therefore the tested
masked diffusion models. Flow matching and continuous score-based text
diffusion require separate evaluation.

\risk{\paragraph{Semantic interpretation.}
Some low-band local-quality effects appear relatively stable, but factuality-related signals are less consistent across datasets and model families. We therefore limit the semantic claim: masking rate changes the reconstruction setting, while local/global or factuality interpretations require task-specific validation.}

\paragraph{Baseline strength.}
The judge comparison is intentionally scoped. Prompted scalar judges can remain
competitive or stronger on narrow scalar tasks, and more optimized judge prompts
could raise the baseline. A point-in-time stronger-judge check showed the same
directional pattern but was not intended as a primary comparison.

\paragraph{Practical cost.}
Band-loss estimates require Monte Carlo samples over both masking rates and
mask patterns, so the method is heavier than AR likelihood or a single prompted
judge call. This cost is acceptable for analysis, but it limits immediate use
as a production metric. Our sample-efficiency controls suggest that the main
SummEval fusion signal survives substantially smaller Monte Carlo budgets, but
estimator optimization remains future work. WMT roundtrip probing is also
compute-asymmetric relative to AR surprisal because it uses repeated
reconstruction across several masking rates.

\section*{Ethical Considerations}

Evaluation metrics influence which generation systems are trusted, iterated on,
and ultimately deployed. In that sense, even a diagnostic method has downstream
impact: emphasizing only a single scalar score can hide failure modes that are
important to users and practitioners. Our goal is to encourage more transparent
evaluation by exposing rate-dependent behavior that scalar metrics may miss.

At the same time, diffusion-based diagnostics inherit biases from the
underlying language models and from the benchmarks used to assess them. A specific misuse risk is treating high-mask behavior as a
guarantee of factual correctness or reliability.
We therefore present the method as a complementary analysis tool and keep it
separate from human oversight or stronger task-specific evaluation protocols.

\bibliography{references}

@inproceedings{bhandari-etal-2020-evaluating,
    title = "Re-evaluating Evaluation in Text Summarization",
    author = "Bhandari, Manik  and
      Gour, Pranav Narayan  and
      Ashfaq, Atabak  and
      Liu, Pengfei  and
      Neubig, Graham",
    editor = "Webber, Bonnie  and
      Cohn, Trevor  and
      He, Yulan  and
      Liu, Yang",
    booktitle = "Proceedings of the 2020 Conference on Empirical Methods in Natural Language Processing (EMNLP)",
    month = nov,
    year = "2020",
    address = "Online",
    publisher = "Association for Computational Linguistics",
    url = "https://aclanthology.org/2020.emnlp-main.751/",
    doi = "10.18653/v1/2020.emnlp-main.751",
    pages = "9347--9359"
}

@inproceedings{kiela-etal-2021-dynabench,
    title = "Dynabench: Rethinking Benchmarking in {NLP}",
    author = "Kiela, Douwe  and
      Bartolo, Max  and
      Nie, Yixin  and
      Kaushik, Divyansh  and
      Geiger, Atticus  and
      Wu, Zhengxuan  and
      Vidgen, Bertie  and
      Prasad, Grusha  and
      Singh, Amanpreet  and
      Ringshia, Pratik  and
      Ma, Zhiyi  and
      Thrush, Tristan  and
      Riedel, Sebastian  and
      Waseem, Zeerak  and
      Stenetorp, Pontus  and
      Jia, Robin  and
      Bansal, Mohit  and
      Potts, Christopher  and
      Williams, Adina",
    editor = "Toutanova, Kristina  and
      Rumshisky, Anna  and
      Zettlemoyer, Luke  and
      Hakkani-Tur, Dilek  and
      Beltagy, Iz  and
      Bethard, Steven  and
      Cotterell, Ryan  and
      Chakraborty, Tanmoy  and
      Zhou, Yichao",
    booktitle = "Proceedings of the 2021 Conference of the North American Chapter of the Association for Computational Linguistics: Human Language Technologies",
    month = jun,
    year = "2021",
    address = "Online",
    publisher = "Association for Computational Linguistics",
    url = "https://aclanthology.org/2021.naacl-main.324/",
    doi = "10.18653/v1/2021.naacl-main.324",
    pages = "4110--4124"
}

@inproceedings{gardner-etal-2020-evaluating,
    title = "Evaluating Models' Local Decision Boundaries via Contrast Sets",
    author = "Gardner, Matt  and
      Artzi, Yoav  and
      Basmov, Victoria  and
      Berant, Jonathan  and
      Bogin, Ben  and
      Chen, Sihao  and
      Dasigi, Pradeep  and
      Dua, Dheeru  and
      Elazar, Yanai  and
      Gottumukkala, Ananth  and
      Gupta, Nitish  and
      Hajishirzi, Hannaneh  and
      Ilharco, Gabriel  and
      Khashabi, Daniel  and
      Lin, Kevin  and
      Liu, Jiangming  and
      Liu, Nelson F.  and
      Mulcaire, Phoebe  and
      Ning, Qiang  and
      Singh, Sameer  and
      Smith, Noah A.  and
      Subramanian, Sanjay  and
      Tsarfaty, Reut  and
      Wallace, Eric  and
      Zhang, Ally  and
      Zhou, Ben",
    editor = "Cohn, Trevor  and
      He, Yulan  and
      Liu, Yang",
    booktitle = "Findings of the Association for Computational Linguistics: EMNLP 2020",
    month = nov,
    year = "2020",
    address = "Online",
    publisher = "Association for Computational Linguistics",
    url = "https://aclanthology.org/2020.findings-emnlp.117/",
    doi = "10.18653/v1/2020.findings-emnlp.117",
    pages = "1307--1323"
}

@inproceedings{ribeiro-etal-2020-beyond,
    title = "Beyond Accuracy: Behavioral Testing of {NLP} Models with {C}heck{L}ist",
    author = "Ribeiro, Marco Tulio  and
      Wu, Tongshuang  and
      Guestrin, Carlos  and
      Singh, Sameer",
    editor = "Jurafsky, Dan  and
      Chai, Joyce  and
      Schluter, Natalie  and
      Tetreault, Joel",
    booktitle = "Proceedings of the 58th Annual Meeting of the Association for Computational Linguistics",
    month = jul,
    year = "2020",
    address = "Online",
    publisher = "Association for Computational Linguistics",
    url = "https://aclanthology.org/2020.acl-main.442/",
    doi = "10.18653/v1/2020.acl-main.442",
    pages = "4902--4912"
}

@inproceedings{pagnoni-etal-2021-understanding,
    title = "Understanding Factuality in Abstractive Summarization with {FRANK}: A Benchmark for Factuality Metrics",
    author = "Pagnoni, Artidoro  and
      Balachandran, Vidhisha  and
      Tsvetkov, Yulia",
    editor = "Toutanova, Kristina  and
      Rumshisky, Anna  and
      Zettlemoyer, Luke  and
      Hakkani-Tur, Dilek  and
      Beltagy, Iz  and
      Bethard, Steven  and
      Cotterell, Ryan  and
      Chakraborty, Tanmoy  and
      Zhou, Yichao",
    booktitle = "Proceedings of the 2021 Conference of the North American Chapter of the Association for Computational Linguistics: Human Language Technologies",
    month = jun,
    year = "2021",
    address = "Online",
    publisher = "Association for Computational Linguistics",
    url = "https://aclanthology.org/2021.naacl-main.383/",
    doi = "10.18653/v1/2021.naacl-main.383",
    pages = "4812--4829"
}

@inproceedings{liu-etal-2023-revisiting,
    title = "Revisiting the Gold Standard: Grounding Summarization Evaluation with Robust Human Evaluation",
    author = "Liu, Yixin  and
      Fabbri, Alex  and
      Liu, Pengfei  and
      Zhao, Yilun  and
      Nan, Linyong  and
      Han, Ruilin  and
      Han, Simeng  and
      Joty, Shafiq  and
      Wu, Chien-Sheng  and
      Xiong, Caiming  and
      Radev, Dragomir",
    editor = "Rogers, Anna  and
      Boyd-Graber, Jordan  and
      Okazaki, Naoaki",
    booktitle = "Proceedings of the 61st Annual Meeting of the Association for Computational Linguistics (Volume 1: Long Papers)",
    month = jul,
    year = "2023",
    address = "Toronto, Canada",
    publisher = "Association for Computational Linguistics",
    url = "https://aclanthology.org/2023.acl-long.228/",
    doi = "10.18653/v1/2023.acl-long.228",
    pages = "4140--4170"
}

@article{fabbri-etal-2021-summeval,
    title = "{S}umm{E}val: Re-evaluating Summarization Evaluation",
    author = "Fabbri, Alexander R.  and
      Kry{\'s}ci{\'n}ski, Wojciech  and
      McCann, Bryan  and
      Xiong, Caiming  and
      Socher, Richard  and
      Radev, Dragomir",
    editor = "Roark, Brian  and
      Nenkova, Ani",
    journal = "Transactions of the Association for Computational Linguistics",
    volume = "9",
    year = "2021",
    address = "Cambridge, MA",
    publisher = "MIT Press",
    url = "https://aclanthology.org/2021.tacl-1.24/",
    doi = "10.1162/tacl_a_00373",
    pages = "391--409"
}

@inproceedings{fu-etal-2024-gptscore,
    title = "{GPTS}core: Evaluate as You Desire",
    author = "Fu, Jinlan  and
      Ng, See-Kiong  and
      Jiang, Zhengbao  and
      Liu, Pengfei",
    editor = "Duh, Kevin  and
      Gomez, Helena  and
      Bethard, Steven",
    booktitle = "Proceedings of the 2024 Conference of the North American Chapter of the Association for Computational Linguistics: Human Language Technologies (Volume 1: Long Papers)",
    month = jun,
    year = "2024",
    address = "Mexico City, Mexico",
    publisher = "Association for Computational Linguistics",
    url = "https://aclanthology.org/2024.naacl-long.365",
    doi = "10.18653/v1/2024.naacl-long.365",
    pages = "6556--6576",
}

@inproceedings{wang-etal-2023-chatgpt,
    title = "Is {C}hat{GPT} a Good {NLG} Evaluator? A Preliminary Study",
    author = "Wang, Jiaan  and
      Liang, Yunlong  and
      Meng, Fandong  and
      Sun, Zengkui  and
      Shi, Haoxiang  and
      Li, Zhixu  and
      Xu, Jinan  and
      Qu, Jianfeng  and
      Zhou, Jie",
    editor = "Dong, Yue  and
      Xiao, Wen  and
      Wang, Lu  and
      Liu, Fei  and
      Carenini, Giuseppe",
    booktitle = "Proceedings of the 4th New Frontiers in Summarization Workshop",
    month = dec,
    year = "2023",
    address = "Singapore",
    publisher = "Association for Computational Linguistics",
    url = "https://aclanthology.org/2023.newsum-1.1",
    doi = "10.18653/v1/2023.newsum-1.1
        
        
        
        ",
    pages = "1--11",
}

@inproceedings{Kocmi2023LargeLM,
  title = "Large Language Models Are State-of-the-Art Evaluators of Translation Quality",
  author = "Kocmi, Tom and Federmann, Christian",
  booktitle = "Proceedings of the 24th Annual Conference of the European Association for Machine Translation",
  month = jun,
  year = "2023",
  address = "Tampere, Finland",
  publisher = "European Association for Machine Translation",
  url = "https://aclanthology.org/2023.eamt-1.19/",
  pages = "193--203"
}

@inproceedings{min-etal-2023-factscore,
    title = "{FA}ct{S}core: Fine-grained Atomic Evaluation of Factual Precision in Long Form Text Generation",
    author = "Min, Sewon  and
      Krishna, Kalpesh  and
      Lyu, Xinxi  and
      Lewis, Mike  and
      Yih, Wen-tau  and
      Koh, Pang  and
      Iyyer, Mohit  and
      Zettlemoyer, Luke  and
      Hajishirzi, Hannaneh",
    editor = "Bouamor, Houda  and
      Pino, Juan  and
      Bali, Kalika",
    booktitle = "Proceedings of the 2023 Conference on Empirical Methods in Natural Language Processing",
    month = dec,
    year = "2023",
    address = "Singapore",
    publisher = "Association for Computational Linguistics",
    url = "https://aclanthology.org/2023.emnlp-main.741/",
    doi = "10.18653/v1/2023.emnlp-main.741",
    pages = "12076--12100"
}

@inproceedings{liu-etal-2023-g,
    title = "{G}-Eval: {NLG} Evaluation using Gpt-4 with Better Human Alignment",
    author = "Liu, Yang  and
      Iter, Dan  and
      Xu, Yichong  and
      Wang, Shuohang  and
      Xu, Ruochen  and
      Zhu, Chenguang",
    editor = "Bouamor, Houda  and
      Pino, Juan  and
      Bali, Kalika",
    booktitle = "Proceedings of the 2023 Conference on Empirical Methods in Natural Language Processing",
    month = dec,
    year = "2023",
    address = "Singapore",
    publisher = "Association for Computational Linguistics",
    url = "https://aclanthology.org/2023.emnlp-main.153",
    doi = "10.18653/v1/2023.emnlp-main.153",
    pages = "2511--2522",
    }

@inproceedings{li2022diffusionlmimprovescontrollabletext,
      title={Diffusion-LM Improves Controllable Text Generation}, 
      author={Xiang Lisa Li and John Thickstun and Ishaan Gulrajani and Percy Liang and Tatsunori B. Hashimoto},
      booktitle={Advances in Neural Information Processing Systems},
      year={2022},
      url={https://openreview.net/forum?id=3s9IrEsjLyk}, 
}

@inproceedings{gong2023diffuseqsequencesequencetext,
      title={DiffuSeq: Sequence to Sequence Text Generation with Diffusion Models}, 
      author={Shansan Gong and Mukai Li and Jiangtao Feng and Zhiyong Wu and Lingpeng Kong},
      booktitle={The Eleventh International Conference on Learning Representations},
      year={2023},
      url={https://openreview.net/forum?id=jQj-_rLVXsj}, 
}

@inproceedings{wu2023ardiffusionautoregressivediffusionmodel,
      title={AR-Diffusion: Auto-Regressive Diffusion Model for Text Generation}, 
      author={Tong Wu and Zhihao Fan and Xiao Liu and Yeyun Gong and Yelong Shen and Jian Jiao and Hai-Tao Zheng and Juntao Li and Zhongyu Wei and Jian Guo and Nan Duan and Weizhu Chen},
      booktitle={Advances in Neural Information Processing Systems},
      year={2023},
      url={https://openreview.net/forum?id=0EG6qUQ4xE}, 
}

@inproceedings{strudel2022selfconditionedembeddingdiffusiontext,
      title={Self-conditioned Embedding Diffusion for Text Generation}, 
      author={Robin Strudel and Corentin Tallec and Florent Altché and Yilun Du and Yaroslav Ganin and Arthur Mensch and Will Grathwohl and Nikolay Savinov and Sander Dieleman and Laurent Sifre and Rémi Leblond},
      booktitle={The Eleventh International Conference on Learning Representations},
      year={2023},
      url={https://openreview.net/forum?id=OpzV3lp3IMC}, 
}

@inproceedings{tang-etal-2023-diffusion,
    title = "Can Diffusion Model Achieve Better Performance in Text Generation? Bridging the Gap between Training and Inference!",
    author = "Tang, Zecheng  and
      Wang, Pinzheng  and
      Zhou, Keyan  and
      Li, Juntao  and
      Cao, Ziqiang  and
      Zhang, Min",
    editor = "Rogers, Anna  and
      Boyd-Graber, Jordan  and
      Okazaki, Naoaki",
    booktitle = "Findings of the Association for Computational Linguistics: ACL 2023",
    month = jul,
    year = "2023",
    address = "Toronto, Canada",
    publisher = "Association for Computational Linguistics",
    url = "https://aclanthology.org/2023.findings-acl.721/",
    doi = "10.18653/v1/2023.findings-acl.721",
    pages = "11359--11386"
}

@inproceedings{zhang2020bertscoreevaluatingtextgeneration,
      title={{BERTScore}: Evaluating Text Generation with {BERT}},
      author={Tianyi Zhang and Varsha Kishore and Felix Wu and Kilian Q. Weinberger and Yoav Artzi},
      booktitle={International Conference on Learning Representations},
      year={2020},
      url={https://openreview.net/forum?id=SkeHuCVFDr},
}

@inproceedings{yuan2021bartscoreevaluatinggeneratedtext,
      title={{BARTScore}: Evaluating Generated Text as Text Generation},
      author={Weizhe Yuan and Graham Neubig and Pengfei Liu},
      booktitle={Advances in Neural Information Processing Systems},
      volume={34},
      year={2021},
      url={https://proceedings.neurips.cc/paper/2021/hash/e4d2b6e6fdeca3e60e0f1a62fee3d9dd-Abstract.html},
}

@inproceedings{qin2022t5scorediscriminativefinetuninggenerative,
    title = "{T5Score}: Discriminative Fine-tuning of Generative Evaluation Metrics",
    author = "Qin, Yiwei  and
      Yuan, Weizhe  and
      Neubig, Graham  and
      Liu, Pengfei",
    booktitle = "Findings of the Association for Computational Linguistics: EMNLP 2023",
    month = dec,
    year = "2023",
    address = "Singapore",
    publisher = "Association for Computational Linguistics",
    url = "https://aclanthology.org/2023.findings-emnlp.1014/",
    pages = "15178--15195"
}

@inproceedings{chiang-lee-2023-large,
    title = "Can Large Language Models Be an Alternative to Human Evaluations?",
    author = "Chiang, Cheng-Han  and
      Lee, Hung-yi",
    editor = "Rogers, Anna  and
      Boyd-Graber, Jordan  and
      Okazaki, Naoaki",
    booktitle = "Proceedings of the 61st Annual Meeting of the Association for Computational Linguistics (Volume 1: Long Papers)",
    month = jul,
    year = "2023",
    address = "Toronto, Canada",
    publisher = "Association for Computational Linguistics",
    url = "https://aclanthology.org/2023.acl-long.870/",
    doi = "10.18653/v1/2023.acl-long.870",
    pages = "15607--15631"
}

@inproceedings{lin-2004-rouge,
    title = "{ROUGE}: A Package for Automatic Evaluation of Summaries",
    author = "Lin, Chin-Yew",
    booktitle = "Text Summarization Branches Out",
    month = jul,
    year = "2004",
    address = "Barcelona, Spain",
    publisher = "Association for Computational Linguistics",
    url = "https://aclanthology.org/W04-1013/",
    pages = "74--81"
}

@inproceedings{papineni-etal-2002-bleu,
    title = "{B}leu: a Method for Automatic Evaluation of Machine Translation",
    author = "Papineni, Kishore  and
      Roukos, Salim  and
      Ward, Todd  and
      Zhu, Wei-Jing",
    editor = "Isabelle, Pierre  and
      Charniak, Eugene  and
      Lin, Dekang",
    booktitle = "Proceedings of the 40th Annual Meeting of the Association for Computational Linguistics",
    month = jul,
    year = "2002",
    address = "Philadelphia, Pennsylvania, USA",
    publisher = "Association for Computational Linguistics",
    url = "https://aclanthology.org/P02-1040/",
    doi = "10.3115/1073083.1073135",
    pages = "311--318"
}

@misc{huang2024ufounifiedflexibleframework,
      title={UFO: a Unified and Flexible Framework for Evaluating Factuality of Large Language Models}, 
      author={Zhaoheng Huang and Zhicheng Dou and Yutao Zhu and {Ji-rong Wen}},
      year={2024},
      eprint={2402.14690},
      archivePrefix={arXiv},
      primaryClass={cs.CL},
      url={https://arxiv.org/abs/2402.14690}, 
}

@inproceedings{nie2025large,
      title={Large Language Diffusion Models},
      author={Shen Nie and Fengqi Zhu and Zebin You and Xiaolu Zhang and Jingyang Ou and Jun Hu and Jun Zhou and Yankai Lin and {Ji-Rong Wen} and Chongxuan Li},
      booktitle={Advances in Neural Information Processing Systems},
      year={2025},
      url={https://openreview.net/forum?id=KnqiC0znVF},
}

@misc{dream2025dream,
      title={Dream 7B: Diffusion Large Language Models},
      author={Jiacheng Ye and Zhihui Xie and Lin Zheng and Jiahui Gao and Zirui Wu and Xin Jiang and Zhenguo Li and Lingpeng Kong},
      year={2025},
      eprint={2508.15487},
      archivePrefix={arXiv},
      primaryClass={cs.CL},
      doi={10.48550/arXiv.2508.15487},
      url={https://arxiv.org/abs/2508.15487},
}

@inproceedings{wang2023assessingreliabilitylargelanguage,
      title={Assessing Factual Reliability of Large Language Model Knowledge}, 
      author={Weixuan Wang and Barry Haddow and Alexandra Birch and Wei Peng},
      booktitle={Proceedings of the 2024 Conference of the North American Chapter of the Association for Computational Linguistics: Human Language Technologies},
      year={2024},
      url={https://aclanthology.org/2024.naacl-long.46/}, 
}

@inproceedings{honovich2022truereevaluatingfactualconsistency,
      title={TRUE: Re-evaluating Factual Consistency Evaluation}, 
      author={Or Honovich and Roee Aharoni and Jonathan Herzig and Hagai Taitelbaum and Doron Kukliansy and Vered Cohen and Thomas Scialom and Idan Szpektor and Avinatan Hassidim and Yossi Matias},
      booktitle={Proceedings of the 2nd Workshop on Document-grounded Dialogue and Conversational Question Answering},
      year={2022},
      url={https://aclanthology.org/2022.dialdoc-1.19/}, 
}

@inproceedings{sellam-etal-2020-bleurt,
    title = "{BLEURT}: Learning Robust Metrics for Text Generation",
    author = "Sellam, Thibault  and
      Das, Dipanjan  and
      Parikh, Ankur",
    editor = "Jurafsky, Dan  and
      Chai, Joyce  and
      Schluter, Natalie  and
      Tetreault, Joel",
    booktitle = "Proceedings of the 58th Annual Meeting of the Association for Computational Linguistics",
    month = jul,
    year = "2020",
    address = "Online",
    publisher = "Association for Computational Linguistics",
    url = "https://aclanthology.org/2020.acl-main.704/",
    doi = "10.18653/v1/2020.acl-main.704",
    pages = "7881--7892"
}

@inproceedings{zhong-etal-2022-towards,
    title = "Towards a Unified Multi-Dimensional Evaluator for Text Generation",
    author = "Zhong, Ming  and
      Liu, Yang  and
      Yin, Da  and
      Mao, Yuning  and
      Jiao, Yizhu  and
      Liu, Pengfei  and
      Zhu, Chenguang  and
      Ji, Heng  and
      Han, Jiawei",
    editor = "Goldberg, Yoav  and
      Kozareva, Zornitsa  and
      Zhang, Yue",
    booktitle = "Proceedings of the 2022 Conference on Empirical Methods in Natural Language Processing",
    month = dec,
    year = "2022",
    address = "Abu Dhabi, United Arab Emirates",
    publisher = "Association for Computational Linguistics",
    url = "https://aclanthology.org/2022.emnlp-main.131/",
    doi = "10.18653/v1/2022.emnlp-main.131",
    pages = "2023--2038"
}

@inproceedings{zhao-etal-2019-moverscore,
    title = "{M}over{S}core: Text Generation Evaluating with Contextualized Embeddings and Earth Mover Distance",
    author = "Zhao, Wei  and
      Peyrard, Maxime  and
      Liu, Fei  and
      Gao, Yang  and
      Meyer, Christian M.  and
      Eger, Steffen",
    editor = "Inui, Kentaro  and
      Jiang, Jing  and
      Ng, Vincent  and
      Wan, Xiaojun",
    booktitle = "Proceedings of the 2019 Conference on Empirical Methods in Natural Language Processing and the 9th International Joint Conference on Natural Language Processing (EMNLP-IJCNLP)",
    month = nov,
    year = "2019",
    address = "Hong Kong, China",
    publisher = "Association for Computational Linguistics",
    url = "https://aclanthology.org/D19-1053/",
    doi = "10.18653/v1/D19-1053",
    pages = "563--578"
}

@misc{yang2025qwen25technicalreport,
  title = "{Qwen2.5} Technical Report",
  author = "Yang, An and Yang, Baosong and Zhang, Beichen and Hui, Binyuan and Zheng, Bo and Yu, Bowen and Li, Chengyuan and Liu, Dayiheng and Huang, Fei and Wei, Haoran and others",
  year = "2025",
  eprint = "2412.15115",
  archivePrefix = "arXiv",
  primaryClass = "cs.CL",
  url = "https://arxiv.org/abs/2412.15115"
}

@inproceedings{conneau-etal-2020-unsupervised,
  title = "Unsupervised Cross-lingual Representation Learning at Scale",
  author = "Conneau, Alexis and Khandelwal, Kartikay and Goyal, Naman and Chaudhary, Vishrav and Wenzek, Guillaume and Guzm{\'a}n, Francisco and Grave, Edouard and Ott, Myle and Zettlemoyer, Luke and Stoyanov, Veselin",
  booktitle = "Proceedings of the 58th Annual Meeting of the Association for Computational Linguistics",
  month = jul,
  year = "2020",
  address = "Online",
  publisher = "Association for Computational Linguistics",
  url = "https://aclanthology.org/2020.acl-main.747/",
  doi = "10.18653/v1/2020.acl-main.747",
  pages = "8440--8451"
}

@inproceedings{freitag-etal-2021-results,
  title = "Results of the {WMT}21 Metrics Shared Task: Evaluating Metrics with Expert-based Human Evaluations on {TED} and News Domain",
  author = "Freitag, Markus and Rei, Ricardo and Mathur, Nitika and Lo, Chi-kiu and Stewart, Craig and Foster, George and Lavie, Alon and Bojar, Ond{\v{r}}ej",
  booktitle = "Proceedings of the Sixth Conference on Machine Translation",
  month = nov,
  year = "2021",
  address = "Online",
  publisher = "Association for Computational Linguistics",
  url = "https://aclanthology.org/2021.wmt-1.73/",
  pages = "733--774"
}

@misc{google-wmt-mqm-human-evaluation,
  title = "Expert-based Human Evaluations for the Submissions of {WMT} 2020, {WMT} 2021, {WMT} 2022 and {WMT} 2023",
  author = "{Google Research}",
  year = "2024",
  howpublished = "GitHub repository",
  url = "https://github.com/google/wmt-mqm-human-evaluation",
  note = "Accessed 2026-05-25"
}

@inproceedings{vasilyev-etal-2020-fill,
  title = "Fill in the {BLANC}: Human-free quality estimation of document summaries",
  author = "Vasilyev, Oleg and Dharnidharka, Vedant and Bohannon, John",
  booktitle = "Proceedings of the First Workshop on Evaluation and Comparison of NLP Systems",
  month = nov,
  year = "2020",
  address = "Online",
  publisher = "Association for Computational Linguistics",
  url = "https://aclanthology.org/2020.eval4nlp-1.2/",
  doi = "10.18653/v1/2020.eval4nlp-1.2",
  pages = "11--20"
}

@inproceedings{salazar-etal-2020-masked,
  title = "Masked Language Model Scoring",
  author = "Salazar, Julian and Liang, Davis and Nguyen, Toan Q. and Kirchhoff, Katrin",
  booktitle = "Proceedings of the 58th Annual Meeting of the Association for Computational Linguistics",
  month = jul,
  year = "2020",
  address = "Online",
  publisher = "Association for Computational Linguistics",
  url = "https://aclanthology.org/2020.acl-main.240/",
  doi = "10.18653/v1/2020.acl-main.240",
  pages = "2699--2712"
}

@misc{openai2026gpt54mini,
  title = "Introducing {GPT-5.4} mini and nano",
  author = "{OpenAI}",
  year = "2026",
  month = mar,
  day = "17",
  url = "https://openai.com/index/introducing-gpt-5-4-mini-and-nano/",
  note = "Accessed 2026-05-25"
}

\appendix
\section{Additional Details}

\subsection{Masked-Diffusion Decomposition}
\label{app:derivation}

We use the absorbing-mask forward process common in masked diffusion language
models. For a sequence $x=(x_1,\ldots,x_N)$, continuous time $t$, and mask
symbol $M$, each token is independently kept with probability $1-\sigma(t)$:
\begin{equation}
\begin{aligned}
q(z_{t,i}\mid x_i) =
&(1-\sigma(t))\mathbf{1}[z_{t,i}=x_i]\\
&+ \sigma(t)\mathbf{1}[z_{t,i}=M].
\end{aligned}
\end{equation}
For this absorbing-state process, the continuous-time variational objective can
be written as
\begin{equation}
\begin{aligned}
\mathcal{L}(x)=
&-\int_0^{1^-}
\frac{\dot{\sigma}(t)}{1-\sigma(t)}
\sum_{i=1}^{N}\\
&\quad
\mathbb{E}_{q(z_t\mid x)}\!\left[
\mathbf{1}[z_{t,i}=M]
\log p_\theta(x_i\mid z_t)
\right]dt .
\end{aligned}
\end{equation}
Changing variables to $s=\sigma(t)$ gives $ds=\dot{\sigma}(t)dt$ and turns the
time integral into a masking-rate integral:
\begin{equation}
\begin{aligned}
\mathcal{L}(x)=
\int_0^{1^-}
\frac{1}{1-s}
\sum_{i=1}^{N}
\mathbb{E}_{m}\left[m_i r_i(x,m)\right]ds,
\end{aligned}
\end{equation}
where $r_i(x,m)=-\log p_\theta(x_i\mid x_{\bar{S}})$ and $S=\{i:m_i=1\}$.
The factor $1/(1-s)$ is the continuous-time absorbing-state weight: as fewer
tokens remain visible, each masked-token reconstruction term receives higher
ELBO weight.

To obtain the band loss used in the main paper, write
\begin{equation}
\sum_i \mathbb{E}[m_i r_i(x,m)]
=
\mathbb{E}\left[
|S|\frac{1}{|S|}\sum_{i\in S}r_i(x,m)
\right].
\end{equation}
More explicitly, letting
$\bar{r}_S(x,m)=\frac{1}{|S|}\sum_{i\in S}r_i(x,m)$ on the event $|S|>0$, the
conditional identity is
\begin{equation}
\begin{aligned}
\mathbb{E}\!\left[|S|\bar{r}_S \mid |S|>0\right]
&=
\mathbb{E}[|S| \mid |S|>0]\ell(s,x) \\
&\quad +
\mathrm{Cov}\!\left(|S|,\bar{r}_S \mid |S|>0\right).
\end{aligned}
\end{equation}
Our estimator replaces the conditional mean mask count by $sN$ and drops the
covariance term, yielding
\begin{equation}
\mathcal{L}(x)\approx
\int_0^{1^-}
\frac{s}{1-s}N\ell(s,x)ds .
\end{equation}
The approximation is therefore neither claimed as an upper bound nor as a lower
bound; it is a pragmatic decomposition whose accuracy depends on both mask-count
concentration and the residual covariance term. The first part is easy to
quantify: for $|S| \sim \mathrm{Binom}(N,s)$, the coefficient of variation is
$\sqrt{(1-s)/(Ns)}$. At the sequence lengths used in the main paper
($N \le 1024$ with most examples near that cap), this gives relative mask-count
fluctuations of about 0.094 at $s=0.1$, 0.031 at $s=0.5$, and 0.010 at
$s=0.9$. What remains uncontrolled is the covariance between mask count and
average masked-token loss. We therefore interpret Eq.~(11) as a useful
decomposition with an explicit omitted term, and we treat exact semantic band
placement as an empirical question.

\paragraph{Empirical magnitude of the omitted covariance.}
To estimate the practical size of this omitted term, we run a small
length-stratified check on a 90-document SummEval subset spanning the observed
summary lengths in that benchmark (21 to 147 tokens). For each document and each
$s \in \{0.1,0.3,0.5,0.7,0.9\}$, we draw 200 independent masks and compute both
the exact Monte Carlo quantity $\mathbb{E}[|S|\bar{r}_S]$ and the approximation
$sN\ell(s,x)$. Table~\ref{tab:covariance-probe} shows that the omitted
covariance is positive in the median, so the approximation underestimates the
exact Monte Carlo term by a small positive bias: the median relative error
ranges from 2.46\% at $s=0.1$ to 0.30\% at $s=0.9$. The empirical mask-count
coefficient of variation closely matches the Binomial prediction at every
tested rate, and the empirical gap between $\mathbb{E}[|S|]$ and $sN$ remains
below 0.3\% in the median across all conditions. Most importantly for the
downstream use in this paper, replacing the exact Monte Carlo term by the
approximation preserves document-level ranking almost perfectly:
Spearman$(\text{exact}, \text{approx}) \ge 0.996$ for all tested masking rates.
The worst-case relative error stays below 10.7\% on this subset.

Within the observed SummEval length range, median relative error decreases
monotonically with document length across all tested masking rates, consistent
with the $|S|$-concentration prediction. A split-half sanity check also shows
stable covariance estimates: the median relative gap between two independent
halves of the 200 mask samples
drops from 2.6\% at $s=0.1$ to 0.05\% at $s=0.9$. We therefore use Eq.~(11) as
an empirically calibrated approximation.

\begin{table}[ht]
\centering
\scriptsize
\setlength{\tabcolsep}{3pt}
\resizebox{.45\textwidth}{!}{
\begin{tabular}{@{}ccccc@{}}
\toprule
$s$ & Med. rel. err. & IQR & Emp./theory CV & Rank $\rho$ \\
\midrule
0.1 & 0.0246 & 0.0317 & 0.344 / 0.338 & 0.9969 \\
0.3 & 0.0207 & 0.0133 & 0.170 / 0.172 & 0.9986 \\
0.5 & 0.0167 & 0.0078 & 0.112 / 0.113 & 0.9996 \\
0.7 & 0.0102 & 0.0053 & 0.073 / 0.074 & 0.9994 \\
0.9 & 0.0030 & 0.0016 & 0.037 / 0.038 & 0.9997 \\
\bottomrule
\end{tabular}}
\caption{Empirical covariance probe for Eq.~(11) on a 90-document SummEval
subset (21 to 147 tokens). Relative error is the signed covariance term divided
by the leading approximation $sN\ell(s,x)$. The empirical mask-count
coefficient of variation closely matches the Binomial prediction, and
document-level rankings induced by the exact and approximate quantities are
nearly identical.}
\label{tab:covariance-probe}
\end{table}

\subsection{SummEval Positioning}
\label{app:summeval-positioning}

Table~\ref{tab:summeval-positioning} places the fixed-evaluator SummEval
fusion results against published scalar references. It is included to calibrate
scope; the main comparison in the paper is residual signal under a fixed
evaluator family.

\begin{table*}[!t]
\centering
\small
\setlength{\tabcolsep}{6pt}
\renewcommand{\arraystretch}{1.14}

\begin{minipage}[t]{0.46\textwidth}
\centering
\begin{tabular}{@{}lcc@{}}
\toprule
\multicolumn{3}{@{}l}{\textbf{A. No-judge signals}} \\
\midrule
\textbf{Signal} & \textbf{Relevance $\rho$} & \textbf{Gain} \\
\midrule
AR likelihood & 0.2106 & n/a \\
Diffusion best band & 0.2209 & n/a \\
AR + full diffusion scalar & 0.2610 & ref. \\
AR + best diffusion band & \textbf{0.2795} & \textbf{+0.0185} \\
\bottomrule
\end{tabular}
\end{minipage}
\hfill
\begin{minipage}[t]{0.46\textwidth}
\centering
\begin{tabular}{@{}lcc@{}}
\toprule
\multicolumn{3}{@{}l}{\textbf{B. Fixed-judge fusion}} \\
\midrule
\textbf{Added signal} & \textbf{Relevance $\rho$} & \textbf{Gain} \\
\midrule
None & 0.3065 & n/a \\
AR & 0.3273 & +0.0208 \\
Diffusion & 0.3454 & +0.0389 \\
AR + diffusion & \textbf{0.3636} & \textbf{+0.0571} \\
\bottomrule
\end{tabular}
\end{minipage}

\caption{
SummEval relevance complementarity under the fixed open-weight setting.
Panel A tests whether preserving the diffusion rate axis helps beyond a
collapsed diffusion scalar; Panel B tests whether diffusion adds residual
signal to the same fixed prompted judge. Published SummEval scalar references
from \citet{liu-etal-2023-g} range from $0.312$ to $0.547$ and are used only
to calibrate scale.
}
\label{tab:summeval-positioning}
\end{table*}

\subsection{Reproducibility and Computational Cost}
\label{app:reproducibility}

The main band-loss configuration uses 10 equal-width bands, 20 masking-rate
samples per band, and 20 mask samples per rate. Thus a single example uses up to
$10\times20\times20=4000$ masked reconstructions before batching. The main
roundtrip configuration uses five masking rates and 100 reconstructions per
rate. For token-level localization, AUROC and thresholded F1 use the conditional
revision frequency in Eq.~\ref{eq:roundtrip}. The implementation also records an
unconditional change-frequency field divided by the total number of
reconstructions, but that diagnostic field is not used for the reported WMT
AUROC or F1 results. All stochastic experiments use seed 13, bfloat16 inference, maximum
sequence length 1024, and single A100-80G jobs unless otherwise stated.

\begin{table*}[t]
\centering
\small
\setlength{\tabcolsep}{4pt}
\begin{tabularx}{\textwidth}{@{}lX@{}}
\toprule
Item & Reproducibility detail \\
\midrule
Code and scripts & The paper source package contains the LaTeX sources,
rendered figures, and rendered tables. Experimental code and configuration files
will be released after de-anonymization. \\
Core models & LLaDA uses \texttt{GSAI-ML/LLaDA-8B-Instruct}; Dream uses
\texttt{Dream-org/Dream-v0-Instruct-7B}; AR and the fixed judge use
\texttt{Qwen/Qwen2.5-7B-Instruct}; the static cloze baseline uses
\texttt{facebook/xlm-roberta-xxl}. \\
Pinned model refs & The local Qwen cache resolves \texttt{main} to
\texttt{a09a35458c702b33eeacc393d103063234e8bc28}; diffusion checkpoints are
recorded by release name in the current artifact and should be archived with
commit hashes before de-anonymized release. \\
Environment & Python 3.11 with package lower bounds
\texttt{torch>=2.1}, \texttt{transformers>=4.46}, \texttt{accelerate>=0.34},
\texttt{datasets>=2.18}, \texttt{scipy>=1.11}, and \texttt{pandas>=2.1};
LLaDA and Dream configs record Transformers 4.46.3 and 4.46.2 respectively. \\
Benchmarks and licenses & SummEval is distributed with
MIT licenses in the local artifact; WMT MQM is distributed under Apache-2.0.
All are used through their released benchmark files and cited in the main text.
\\
WMT preprocessing & The loader keeps \texttt{hyp.*} systems from
newstest2021 zh-en, groups rows by system/document/segment, strips MQM markup,
projects spans to character-level masks, derives severity slices from released
segment-level counts, and aggregates rater penalties with the released severity
weights. \\
Randomness and hardware & Seed 13, bfloat16 inference, max length 1024, and
single A100-80G jobs unless otherwise stated; WMT significance uses 2,000
segment-level bootstrap resamples with paired comparisons matched by segment
id.
\\
\bottomrule
\end{tabularx}
\caption{Reproducibility summary for the main experiments. The table separates
what is already fixed in the anonymous artifact from what should be pinned
again before public release, especially diffusion checkpoint commit hashes.}
\label{tab:reproducibility-summary}
\end{table*}

The cost is substantial. On our cluster, Qwen AR scoring on SummEval completed
in 4 minutes 54 seconds, while LLaDA band-loss scoring on the same 1,600
examples took 2 days 23 hours 21 minutes under the high-fidelity Monte Carlo
configuration. WMT MQM full roundtrip over 8,431 segments took 5 hours 10
minutes. These numbers motivate our positioning: masked-diffusion scoring is an
analysis tool for exposing diagnostic structure and should be treated
separately from low-cost AR likelihood.

Our prompted judges use the same instruction: ``Rate only whether the summary
covers the important content of the source article.'' The model receives the
source article and summary, then must return only JSON of the form
\texttt{\{"score": 4\}}, where the score is an integer from 1 to 5. We parse
only this JSON score and discard unparsable outputs. The Qwen judge is
\texttt{Qwen2.5-7B-Instruct}; the stronger-judge check uses \texttt{gpt-5.4-mini} \cite{openai2026gpt54mini} through an API endpoint in April 2026 with deterministic decoding. Because commercial API models may change over time, we treat this experiment as a point-in-time robustness check. The static bidirectional cloze baseline uses
\texttt{facebook/xlm-roberta-xxl}.

Fusion uses a grid over $\alpha\in\{0,0.1,\ldots,1\}$ after z-scoring the
diffusion and AR signals. To avoid interpreting test-set tuning as a deployed
metric, we also run 5-fold cross-validation on SummEval grouped by source
document: all system outputs for the same article remain in the same fold. For
each fold, the diffusion band and fusion weight are selected on the training
split and evaluated on the held-out split. For relevance, grouped
cross-validated fusion reaches $\rho=0.2795$, close to the full-data diagnostic
value of 0.2857; all folds select band 8 and weights near $\alpha=0.5$.

\begin{table}[t]
\centering
\scriptsize
\setlength{\tabcolsep}{3pt}
\resizebox{\linewidth}{!}{
\begin{tabular}{@{}llccccc@{}}
\toprule
Model & Dim. & Band & AR & Band $\rho$ & AR+band & Gain \\
\midrule
\multirow{4}{*}{LLaDA}
& Rel. & 8 & 0.2106 & 0.2209 & 0.2857 & +0.0648 \\
& Cons. & 7 & 0.1022 & 0.1592 & 0.1849 & +0.0257 \\
& Flu. & 0 & 0.1699 & 0.1709 & 0.1887 & +0.0177 \\
& Coh. & 1 & 0.1695 & 0.0869 & 0.1695 & +0.0000 \\
\midrule
\multirow{4}{*}{Dream}
& Rel. & 9 & 0.2106 & 0.2024 & 0.2638 & +0.0532 \\
& Cons. & 8 & 0.1022 & 0.1239 & 0.1465 & +0.0226 \\
& Flu. & 0 & 0.1699 & 0.0287 & 0.1794 & +0.0095 \\
& Coh. & 0 & 0.1695 & 0.0388 & 0.1784 & +0.0088 \\
\bottomrule
\end{tabular}}
\caption{Full-data SummEval per-dimension AR and diffusion-band
complementarity. Band correlations are sign-aligned Spearman correlations for
the selected diffusion band. Gain is AR+band minus the stronger single signal.
This table supplies the detailed relevance, consistency, fluency, and coherence
breakdown summarized in Section~5.1.}
\label{tab:summeval-dimension-breakdown}
\end{table}

\begin{table}[t]
\centering
\scriptsize
\setlength{\tabcolsep}{3pt}
\resizebox{\linewidth}{!}{
\begin{tabular}{@{}llccccc@{}}
\toprule
Model & Dim. & AR & AR+scalar & AR+band & $\Delta$ & $p$ \\
\midrule
\multirow{4}{*}{LLaDA}
& Rel. & 0.2106 & 0.2610 & 0.2795 & +0.0185 & .0015 \\
& Cons. & 0.1022 & 0.1428 & 0.1705 & +0.0276 & .0085 \\
& Flu. & 0.1699 & 0.1692 & 0.1852 & +0.0160 & .0895 \\
& Coh. & 0.1695 & 0.1590 & 0.1530 & -0.0059 & .9675 \\
\midrule
\multirow{4}{*}{Dream}
& Rel. & 0.2106 & 0.2524 & 0.2517 & -0.0006 & .6577 \\
& Cons. & 0.1022 & 0.1361 & 0.1423 & +0.0062 & .0055 \\
& Flu. & 0.1699 & 0.1692 & 0.1693 & +0.0001 & .4848 \\
& Coh. & 0.1695 & 0.1609 & 0.1517 & -0.0094 & .8161 \\
\bottomrule
\end{tabular}}
\caption{SummEval scalarization controls. AR+scalar fuses AR likelihood with
the full diffusion scalar from the same diffusion model, while AR+band fuses AR
with the best masking-rate band selected within grouped-CV folds. $\Delta$ is
AR+band minus AR+scalar.}
\label{tab:summeval-scalarization-controls}
\end{table}

\subsection{Efficiency and Sensitivity Controls}
\label{app:efficiency-sensitivity}

Table~\ref{tab:mc-budget} reports a direct Monte Carlo budget check on
SummEval relevance. Reducing the estimator from $20\times20$ masking-rate and
mask samples to $5\times5$ preserves the main diffusion and AR fusion result,
while reducing runtime from the original high-fidelity multi-day run to under
four hours on the same cluster setting. The complementarity signal persists
with far fewer than 4,000 reconstructions per example.

\begin{table}[ht]
\centering
\small
\setlength{\tabcolsep}{4pt}
\begin{tabularx}{\linewidth}{@{}lcccc@{}}
\toprule
Budget & Runtime & Band $\rho$ & Fusion $\rho$ & Gain \\
\midrule
$5\times5$ & 3h43m & 0.2263 & 0.2837 & 0.0574 \\
$10\times10$ & 9h06m & 0.2204 & 0.2786 & 0.0582 \\
$20\times20$ & 2d23h21m & 0.2209 & 0.2795 & 0.0586 \\
\bottomrule
\end{tabularx}
\caption{Monte Carlo budget sensitivity on SummEval relevance. The
$20\times20$ runtime is the original high-fidelity run; lower-budget runs were
executed separately and preserve the same fusion gain.}
\label{tab:mc-budget}
\end{table}

Table~\ref{tab:band-boundary} checks whether the relevance finding depends on
the arbitrary choice of 10 equal-width bands. The exact selected index changes
with the discretization, but cross-validated fusion remains positive and the
selected region stays in the high-mask part of the axis.

\begin{table}[ht]
\centering
\small
\setlength{\tabcolsep}{4pt}
\begin{tabularx}{\linewidth}{@{}lcccX@{}}
\toprule
Bands & Band $\rho$ & Fusion $\rho$ & Gain & Selected region \\
\midrule
5 & 0.1668 & 0.2521 & 0.0414 & mostly band 4 \\
10 & 0.2209 & 0.2795 & 0.0586 & band 8 in all folds \\
20 & 0.1799 & 0.2778 & 0.0672 & bands 16 and 17 \\
\bottomrule
\end{tabularx}
\caption{Band-boundary sensitivity for SummEval relevance. The high-mask
region remains selected under 5, 10, and 20 equal-width partitions.}
\label{tab:band-boundary}
\end{table}

Table~\ref{tab:sample-shuffle-control} breaks the rate order while preserving
the LLaDA reconstruction sample budget. For each SummEval example, the 200
stored reconstruction samples are randomly repartitioned into ten pseudo-bands.
The same grouped-CV selection and fusion protocol is then applied to each
shuffled run.

\begin{table}[ht]
\centering
\small
\setlength{\tabcolsep}{4pt}
\begin{tabularx}{\linewidth}{@{}l c >{\raggedright\arraybackslash}X@{}}
\toprule
Signal structure & AR fusion $\rho$ & Notes \\
\midrule
Observed rate bands & 0.2795 & grouped-CV result \\
Full diffusion scalar & 0.2610 & scalarized LLaDA \\
Shuffled pseudo-bands & 0.2098 & mean over 100 shuffles \\
Best shuffled run & 0.2431 & maximum over 100 shuffles \\
\bottomrule
\end{tabularx}
\caption{Sample-shuffle control for SummEval relevance. Shuffled pseudo-bands
preserve the number of reconstruction samples while removing the masking-rate
ordering; the 95\% shuffle range for AR+pseudo-band fusion is
$[0.1911,0.2362]$.}
\label{tab:sample-shuffle-control}
\end{table}

Table~\ref{tab:wmt-ar-diffusion-tradeoff} summarizes the WMT operating-point
comparison discussed in Section~\ref{sec:wmt-localization}.

\begin{table}[t]
\centering
\small
\setlength{\tabcolsep}{4pt}
\begin{tabular}{@{}lccc@{}}
\toprule
Metric & Diffusion best & AR & Winner \\
\midrule
Overall AUROC & 0.6705 & 0.6330 & Diff. \\
Pooled F1 & 0.2311 & 0.2636 & AR \\
Major F1 & 0.3464 & 0.2811 & Diff. \\
Minor F1 & 0.2852 & 0.2115 & Diff. \\
\bottomrule
\end{tabular}
\caption{WMT localization comparison at the reported operating points. Results
are mixed: diffusion leads on ranking AUROC and severity-specific F1, while AR
leads on pooled fixed-threshold F1.}
\label{tab:wmt-ar-diffusion-tradeoff}
\end{table}

Table~\ref{tab:threshold-sensitivity} reports cross-fitted threshold selection
for WMT MQM token-level F1. Thresholds are selected on held-out folds and
evaluated on separate folds. Diffusion remains ahead of both AR surprisal and
the XLM-R bidirectional cloze baseline on macro-F1 for error-containing,
major-error, and minor-only slices.

\begin{table}[ht]
\centering
\scriptsize
\setlength{\tabcolsep}{3pt}
\resizebox{.35\textwidth}{!}{
\begin{tabular}{@{}lcccc@{}}
\toprule
Slice & Diff. .3 & Diff. .5 & AR & XLM-R \\
\midrule
Err. & 0.3251 & 0.3246 & 0.2987 & 0.2874 \\
Major & 0.3385 & 0.3381 & 0.3116 & 0.3018 \\
Minor & 0.2832 & 0.2838 & 0.2602 & 0.2443 \\
\bottomrule
\end{tabular}}
\caption{Cross-fitted macro-F1 under threshold sensitivity analysis on WMT
MQM. Diff. .3 and Diff. .5 denote diffusion at $s=0.3$ and $s=0.5$; Err.
denotes error-containing segments. Threshold selection leaves the comparison
with AR and the bidirectional cloze baseline unchanged.}
\label{tab:threshold-sensitivity}
\end{table}

\subsection{Inductive-Bias Slices}
\label{app:inductive-bias-slices}

Table~\ref{tab:wmt-position-slice} groups WMT MQM segments by where the gold
error span falls in the hypothesis. This is a descriptive stress test for the
localization claim. If the roundtrip gain were mainly an artifact of one
position bucket, the advantage over AR token surprisal would likely collapse in at
least one bucket. Instead, LLaDA stays ahead for beginning, middle, and end
errors. The gap is largest for end-position errors, but we treat this as a
descriptive pattern; position alone fails to explain the roundtrip advantage.

\begin{table}[t]
\centering
\scriptsize
\setlength{\tabcolsep}{2.5pt}
\resizebox{.45\textwidth}{!}{
\begin{tabular}{@{}lccccc@{}}
\toprule
Pos. & Rec. & Diff. AUC & AR AUC & Diff. F1 & AR F1 \\
\midrule
Beginning & 1,215 & 0.6587 & 0.6053 & 0.2763 & 0.2344 \\
Middle & 3,168 & 0.6365 & 0.5579 & 0.3582 & 0.2437 \\
End & 1,504 & 0.6478 & 0.5263 & 0.2809 & 0.1652 \\
\bottomrule
\end{tabular}}
\caption{WMT MQM error-position stress test. Diffusion AUROC uses roundtrip
scores at $s=0.5$; diffusion F1 uses $s=0.3$. AR F1 uses the same per-segment
top-20\% token-surprisal rule as the AR token baseline.}
\label{tab:wmt-position-slice}
\end{table}

Table~\ref{tab:summeval-case-slices} gives three qualitative SummEval
relevance slices selected by rank-error agreement among the prompted judge, AR
likelihood, and diffusion. These cases illustrate the inductive-bias
interpretation; the quantitative tables carry the main evidence. The
judge+AR case is a high-quality factual news summary that diffusion ranks too
low. The judge+diffusion case is locally repetitive and off-quality; AR ranks
it differently from the human ranking, while the judge and diffusion assign it
near-human rank. The AR+diffusion case is broadly on-topic and informative, but
the judge assigns a noticeably lower rank than the human ordering. This is the
one place where we explicitly surface a judge-side hypothesis: prompted judges
may prefer more elaborated summaries, so a terse factual summary can be
penalized even when probability-based signals stay close to the human order.


\newcommand{\rankerrors}[3]{%
{\footnotesize
Judge: #1 \quad AR: #2 \quad Diff.: #3
}%
}


\newcommand{\sliceinfo}[4]{%
\textbf{#1}\par
\vspace{4pt}
{\scriptsize
\setlength{\tabcolsep}{3pt}
\begin{tabular}{@{}lccc@{}}
\emph{rank err.}$\downarrow$ & J & AR & Diff. \\
& #2 & #3 & #4
\end{tabular}
}
}

\newcommand{\caseinfo}[2]{%
\textbf{#1}\par
\vspace{3pt}
#2
}




\begin{table*}[t]
\centering
\footnotesize
\setlength{\tabcolsep}{0pt}
\renewcommand{\arraystretch}{1.12}

\begin{tabularx}{\textwidth}{
@{}
>{\raggedright\arraybackslash}p{0.30\textwidth}
@{\hspace{0.045\textwidth}}
>{\raggedright\arraybackslash}X
@{}}
\toprule
\textbf{Slice} & \textbf{Failure mode and excerpt} \\
\midrule

\textbf{Judge + AR}\par
{\footnotesize\emph{w/o Diff.}}\par
\vspace{4pt}
{\scriptsize
\setlength{\tabcolsep}{3pt}
\begin{tabular}{@{}lccc@{}}
\emph{rank err.}$\downarrow$ & J & AR & Diff. \\
& 73.5 & 29.0 & 1329.0
\end{tabular}
}
&
\textbf{Long-range ordering sensitivity}\par
\vspace{3pt}
Eight Iranian border guards have been killed in clashes with militants near the
border with Pakistan. A militant group called Jaish al Adal claimed
responsibility for the attack.
\\

\addlinespace[8pt]

\textbf{Judge + Diff.}\par
{\footnotesize\emph{w/o AR}}\par
\vspace{4pt}
{\scriptsize
\setlength{\tabcolsep}{3pt}
\begin{tabular}{@{}lccc@{}}
\emph{rank err.}$\downarrow$ & J & AR & Diff. \\
& 15.0 & 1513.0 & 3.0
\end{tabular}
}
&
\textbf{Local fluency trap under repetition}\par
\vspace{3pt}
Liverpool beat Newcastle United 2 to 0. ``I thought it was a penalty,'' Allen said.
``I thought it was a penalty. I thought it was a penalty.''
\\

\addlinespace[8pt]

\textbf{AR + Diff.}\par
{\footnotesize\emph{w/o Judge}}\par
\vspace{4pt}
{\scriptsize
\setlength{\tabcolsep}{3pt}
\begin{tabular}{@{}lccc@{}}
\emph{rank err.}$\downarrow$ & J & AR & Diff. \\
& 1425.5 & 32.5 & 181.5
\end{tabular}
}
&
\textbf{Possible judge preference mismatch for terse factual summaries}\par
\vspace{3pt}
Wales' crunch Euro 2016 qualifier with Belgium has been declared a 33,000
sell-out. The top two sides in Group B meet in Cardiff on June 12, with Wales
in their best position to qualify for a major tournament since 1958.
\\

\bottomrule
\end{tabularx}

\caption{
Qualitative SummEval relevance slices. Each row shows a case where one
evaluator disagrees with the other two; rank errors are measured against the
human ranking, with lower values better.
}
\label{tab:summeval-case-slices}
\end{table*}

\subsection{Representation Probe}
\label{app:rep-probe}

Table~\ref{tab:representation-probe} reports a small hidden-state probing
sanity check on LLaDA roundtrip states for WMT MQM. We extract masked-token
hidden states at $s\in\{0.1,0.5,0.9\}$ and train grouped-split probes, so the
reported values are held-out values.
Token-surface labels are local lexical classes of the masked token, while
major-error labels are segment-level MQM indicators predicted from mean
masked-token states. The main value of this table is robustness of ordering:
both linear and nonlinear probes preserve the same broad pattern that low-mask
states are easier to decode than high-mask states. The linear probe supports
the local side of the information-resolution interpretation, since token-surface
information is much more linearly decodable at low masking than at high
masking. A nonlinear MLP probe raises major-error AUROC at $s=0.9$ from 0.523
to 0.568, but it also raises low-mask decoding from 0.659 to 0.737 at
$s=0.1$. We therefore avoid reading this as evidence that high-mask states have a
special nonlinear privilege. Instead, we take it as a calibration check that
the mask-rate ordering remains stable under a stronger decoder. Stronger
mechanistic evidence would require nonlinear probes beyond this small check, or
causal methods such as activation patching and controlled representation
interventions.

\begin{table}[ht]
\centering
\small
\setlength{\tabcolsep}{4pt}
\begin{tabular}{@{}lccc@{}}
\toprule
Probe & $s=0.1$ & $s=0.5$ & $s=0.9$ \\
\midrule
Token surface F1 (linear) & 0.830 & 0.611 & 0.244 \\
Major-error AUROC (linear) & 0.659 & 0.652 & 0.523 \\
Major-error F1 (linear) & 0.567 & 0.546 & 0.479 \\
Token surface F1 (MLP) & 0.849 & 0.638 & 0.331 \\
Major-error AUROC (MLP) & 0.737 & 0.691 & 0.568 \\
Major-error F1 (MLP) & 0.685 & 0.592 & 0.570 \\
\bottomrule
\end{tabular}
\caption{Hidden-state probing sanity check on WMT MQM. Low masking preserves
more decodable local token-surface information. Nonlinear probes partially
recover major-error signal at all mask rates, but the low-to-high decodability
ordering remains intact, so this table is best read as an ordering-robustness
check without claiming a white-box proof of a strict information-resolution
law.}
\label{tab:representation-probe}
\end{table}

\paragraph{Exploratory patching check.}
As a minimal follow-up, we patch low-mask hidden states into high-mask forward
passes on a small WMT MQM subset. The check is qualitative and underpowered for
effect-size claims, so we use it only as a design note: reconstruction loss
moves differently under structured patching than under random patching. It
leaves open whether higher-mask states encode globally distinct error semantics.

\subsection{Dream WMT Replication}
\label{app:dream-wmt}

Table~\ref{tab:dream-wmt} reports the same WMT MQM roundtrip aggregation with
Dream-7B-Instruct. This is a negative replication of the strongest LLaDA
localization result: Dream retains a masking-rate-dependent curve, but the absolute
AUROC values are much lower and the best masking rates shift upward. We
therefore treat WMT localization as strong evidence for LLaDA and as an
explicit model-dependence boundary for masked diffusion evaluation.

\begin{table}[ht]
\centering
\scriptsize
\setlength{\tabcolsep}{3pt}
\resizebox{!}{.05\textwidth}{
\begin{tabular}{@{}lcccc@{}}
\toprule
Model & Maj. AUC & Maj. F1 & Min. AUC & Min. F1 \\
\midrule
LLaDA & 0.6774 @ .5 & 0.3464 @ .3 & 0.6506 @ .5 & 0.2852 @ .3 \\
Dream & 0.5871 @ .7 & 0.3134 @ .9 & 0.5418 @ .7 & 0.2597 @ .9 \\
\bottomrule
\end{tabular}}
\caption{Dream replication on WMT MQM zh-en. Values report each model's best
masking rate for the metric. Dream is weaker and shifts toward higher masking rates.}
\label{tab:dream-wmt}
\end{table}



\end{document}